\documentclass[conference]{IEEEtran}
\IEEEoverridecommandlockouts
\usepackage{cite}
\usepackage{amsmath,amssymb,amsfonts}
\usepackage{graphicx}
\usepackage{textcomp}
\usepackage{xcolor}
\usepackage{tabularx}
\usepackage{lscape}
\usepackage{multirow}
\usepackage{booktabs}
\usepackage{hyperref}
\usepackage{subfiles}
\usepackage{dsfont}
\usepackage{lipsum}

\usepackage[super]{nth}
\def\BibTeX{{\rm B\kern-.05em{\sc i\kern-.025em b}\kern-.08em
    T\kern-.1667em\lower.7ex\hbox{E}\kern-.125emX}}
\begin{document}

\title{
Sparse Multitask Learning for Efficient Neural Representation of Motor Imagery and Execution
\footnote{{\thanks{This work was supported by the Institute of Information \& Communications Technology Planning \& Evaluation (IITP) grant, funded by the Korea government (MSIT) (No. 2019-0-00079, Artificial Intelligence Graduate School Program (Korea University); No.2021-0-02068, Artificial Intelligence Innovation Hub) and the National Research Foundation of Korea (NRF) grant funded by the MSIT (No.2022-2-00975, MetaSkin: Developing Next-generation Neurohaptic Interface Techonology that enables Communication and Control in Metaverse by Skin Touch.}
}}
}

\author{\IEEEauthorblockN{Hye-Bin Shin}
\IEEEauthorblockA{\textit{Dept. of Brain and Cognitive Engineering} \\
\textit{Korea University}\\
Seoul, Republic of Korea \\
hb\_shin@korea.ac.kr}\\
\and
\IEEEauthorblockN{Kang Yin}
\IEEEauthorblockA{\textit{Dept. of Artificial Intelligence} \\
\textit{Korea University} \\
Seoul, Republic of Korea \\
charles\_kang@korea.ac.kr} \\
\and
\IEEEauthorblockN{Seong-Whan Lee}
\IEEEauthorblockA{\textit{Dept. of Artificial Intelligence}\\
\textit{Korea University} \\
Seoul, Republic of Korea \\
sw.lee@korea.ac.kr}\\
}

\maketitle

\begin{abstract}
In the quest for efficient neural network models for neural data interpretation and user intent classification in brain-computer interfaces (BCIs), learning meaningful sparse representations of the underlying neural subspaces is crucial. The present study introduces a sparse multitask learning framework for motor imagery (MI) and motor execution (ME) tasks, inspired by the natural partitioning of associated neural subspaces observed in the human brain. Given a dual-task CNN model for MI-ME classification, we apply a saliency-based sparsification approach to prune superfluous connections and reinforce those that show high importance in both tasks. Through our approach, we seek to elucidate the distinct and common neural ensembles associated with each task, employing principled sparsification techniques to eliminate redundant connections and boost the fidelity of neural signal decoding. Our results indicate that this tailored sparsity can mitigate the overfitting problem and improve the test performance with small amount of data, suggesting a viable path forward for computationally efficient and robust BCI systems.
\end{abstract}

\begin{small}
\end{small}

\section{INTRODUCTION}
The neural processes underlying imagination, observation, and execution of movements share a profound and enigmatic connection. Neuroscientific studies have consistently demonstrated that these processes engage overlapping neural circuits, suggesting a shared neural subspace\cite{mimemo2018, structure, manifolds2017Neuron}. This revelation has significant implications for EEG-based brain-computer interfaces (BCIs) leveraging motor imagery (MI) and motor execution (ME) paradigms\cite{intuitiveMI, inter-taskTNSRE, review_paradigms}. Despite the success of machine learning algorithms in uncovering meaningful and discriminative features from EEG data\cite{sst-features-mi, emotions, denoising-matrix}, deep learning has gained popularity in the EEG-BCI domain due to its ability to provide end-to-end learning solutions\cite{review_dl, bangNNLS-sst}. However, state-of-the-art deep learning models trained offline on a fixed EEG dataset show severe overfitting due to the limited amount of training samples, coupled with the high-dimensional nature of multi-channel EEG data\cite{review_dl}. 

To enhance generalization in deep learning models for BCI, various strategies have been employed, from multi-domain learning, which leverages insights gained from multiple subjects data, to multi-modal and multi-view learning approaches that integrate diverse physiological signals and feature representations, thereby enriching the model's understanding and representational power\cite{kim2022dg, gitgan, subject-transfer, 11intuitive, fbcnet}.
Multi-task learning, on the other hand, remains relatively underexplored in the EEG-BCI domain and and presents a unique opportunity to address the challenge of model generalization, especially given its recent success in other domains, e.g., training large language models\cite{mtl_llm2021}.
In the context of motor-related BCI paradigms, a multitask setup is particularly pertinent as it can exploit the duality of task imagination and execution inherent in their design, e.g., motor imagery/execution\cite{11intuitive} and imagined/overt speech\cite{imagined1, imagined2}. Given the neuroscientific evidence for shared neural substrates underlying these tasks, multitask learning can potentially uncover the commonalities and learn shared representation across these related tasks. 

In this study, we construct a multitask CNN model for MI and ME classification and apply sparse training technique to distill the network to its most informative parameters while pruning away those that are not helpful for either task\cite{kd-ref, pruning-ref, sparse-ref}. This sparse multitask learning approach can mitigate model overfitting and improve generalization performance by learning shared representations across different tasks. The integration of different pruning and sparse training methods is largely unexplored in BCI, but it can potentially help to create efficient and compact models suitable for real-time applications\cite{brain-inspired-pruning, kim2023calibration, sensory-response, sparse-ref2}. 
Our contribution is twofold: 
\begin{itemize}
    \item We demonstrate how multitask learning can be effectively applied to motor imagery and execution tasks in BCIs, uncovering shared neural representations that enhance model generalization.
    \item We explore the integration of neural network pruning within this multitask framework, showcasing its potential in reducing model complexity while maintaining, or even improving, performance.
\end{itemize}

In short, this work aims to address the problem of limited generalization performance in deep learning-based BCIs by integrating multitask learning with model pruning techniques. Importantly, we base our construction on the core assumption of shared neural substrates between imagery and execution in the human brain and leverage the dual-task structure common in many motor-related BCI paradigms. 

\section{METHDOLOGY}

\begin{figure}[!t]
\centerline{\includegraphics[width=0.85\columnwidth]{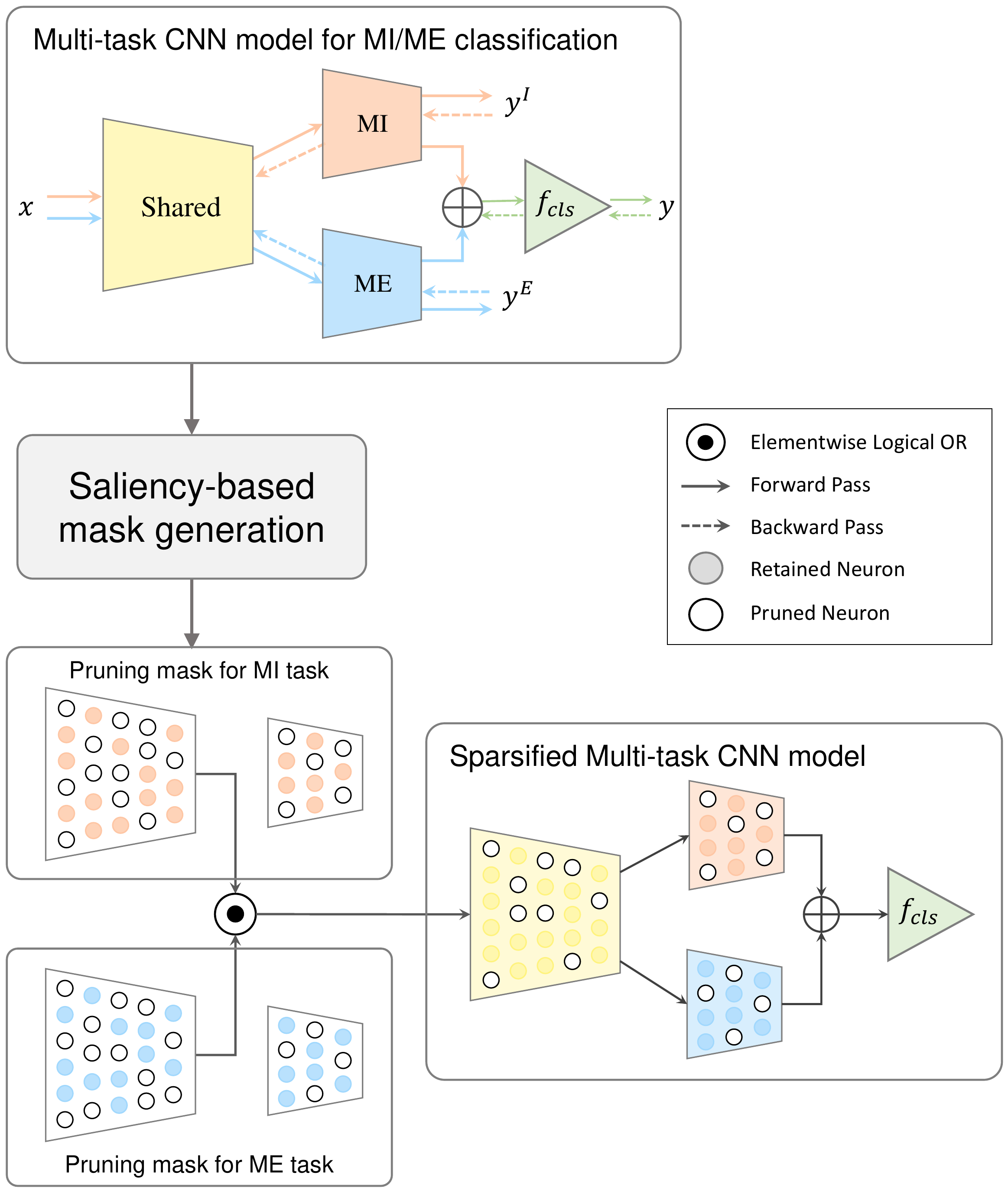}}
\caption{Overview of the proposed sparse multitask learning framework based on static sparse training or ``pruning at initialization." Given a multitask CNN model for MI and ME classification, comprising feature extraction modules and a classification head, the parameters for the shared and private modules undergo a saliency-based mask generation process that outputs a pruning mask for each task. The mask for the shared part is generated by feeding the task-specific masks into an elementwise arbiter function. The final sparsified model is obtained by applying these masks to the respective network parameters.}
\label{fig1}
\end{figure}

\subsection{Problem Statement}
Given $N$ trials of EEG samples $\mathbf{x}_i\in\mathbb{R}^{E\times{T}}$, corresponding labels $y_{i}$, and target sparsity level \(\sigma \in (0,1)\), the following pruning objective can be formulated: 
\begin{equation}
\begin{aligned}
& \min _{\phi} \mathcal{L}(f_\phi, \mathcal{D})=\min _{\phi} \frac{1}{N} \sum_{i=1}^{N} \ell\left(f_\phi\left(x_{i}\right), y_{i}\right) \\
& \text { s.t. } \quad \phi \in \mathbb{R}^{d}, \quad\|\phi\|_{0} \leq(1-\sigma) \cdot d,
\end{aligned}
\end{equation}
where \(\ell(\cdot)\) denotes the pre-defined loss function, \(\phi\) represents the network parameters, \(d\) is the total number of parameters, and \(\|\cdot\|_{0}\) indicates the \(L_{0}\) norm. The problem can be reformulated in terms of finding an optimal binary mask \(\mathcal{M}\):
\begin{equation}
\begin{aligned}
& \min _{\mathcal{M}} \mathcal{L}(f_\phi, \mathcal{M}, \mathcal{D}) =\min _{\mathcal{M}} \frac{1}{N} \sum_{i=1}^{N}  \ell\left(f\left(x_{i} ; \phi \odot \mathcal{M} \right), y_{i}\right) \\
& \text { s.t. } \quad \mathcal{M} \in \mathbb{R}^{d}, \mathcal{M} \in\{0,1\}^{d},\|\mathcal{M}\|_{0} \leq(1-\sigma) \cdot d.
\end{aligned}
\end{equation}

In the context of disentangling the subspaces for imagery and execution, we partition the network parameters into distinct sets of parameters: \(\phi^{\mathcal{I}}\) and \(\phi^{\mathcal{E}}\), for task-specific parameters, and \(\phi^{s}\) for shared parameters. The corresponding binary masks \(\mathcal{M}^{\mathcal{I}}\), \(\mathcal{M}^{\mathcal{E}}\), and \(\mathcal{M}^{s}\) are then optimized to selectively activate the network for each subspace under the sparsity constraint.
The cumulative loss for a sample \(x_{i}\) is a summation over the losses corresponding to each subspace:
\begin{equation}
\ell(f_\phi(x_{i}), y_{i}) = \sum_{k \in \{\mathcal{I}, \mathcal{E}\}} \lambda^{k} \ell^{k}(f(x_{i} ; \phi^{ks} \odot \mathcal{M}^{ks}), y_{i}^{k}),
\end{equation}
where \(\lambda^{k}\) denotes the regularization weight for the loss associated with each subspace and $\phi^{ks} = \phi^{k} \cup \phi^{s}$. The sparsely activated parameters for a task are thus \(\phi^{ks} \odot \mathcal{M}^{ks}\).

\subsection{Proposed Framework}
Our sparse multitask learning framework focuses on optimizing the binary masks \(\mathcal{M}^{\mathcal{I}}\) and \(\mathcal{M}^{\mathcal{E}}\) for the imagery and execution tasks, alongside the shared parameter mask \(\mathcal{M}^{s}\), at the beginning of training. 
Specifically, this initial phase of static sparse training involves evaluating the importance or saliency of each network parameter for the specific tasks of MI and ME classification and pruning the network accordingly. 

For computing the saliency scores, we measured the sensitivity of the loss function to the removal of each parameter, following the sensitivity-based one-shot evaluation of network parameters introduced in SNIP\cite{snip}.  Sensitivity is determined based on the gradient of the loss function with respect to each parameter, indicating how much a small change in the weight would affect the loss, and parameters with higher sensitivity scores are considered more crucial for each task. Based on the computed scores, we can generate the task-specific binary mask $\mathcal{M}^{k}$ for $k \in \{\mathcal{I}, \mathcal{E}\}$ for selecting only the top-ranked parameters. Given the target sparsity level $\sigma$, we want to retain the top $(1-\sigma) \cdot d^{ks}$ parameters, where $d^{ks}$ is the number of task-specific network parameters. 
These masks are used to prune the network, deactivating less critical parameters and enforcing the target sparsity level. 

The next crucial step is the integration of these task-specific masks to form the final mask for the shared parameters, $\mathcal{M}^{s}$. 
The individual masks are input into an arbiter function $\mathcal{A}$, or an element-wise logical OR, which integrates the importance assessments from all tasks and ensures that only universally non-critical parameters are pruned. The efficacy of this approach hinges on the careful balance between task specificity and commonality, enabling our multitask learning framework to specialize and generalize where necessary.
Finally, the network is initialized and trained with these static masks in place, such that only the parameters indicated by the masks are updated during training. The loss for each task is calculated as previously defined. 
The resulting sparse multi-task CNN model was evaluated on a validation set to assess its performance in both MI and ME classification tasks. 

\section{EXPERIMENTS}
\subsection{Dataset Description}
The study utilized a publicly available dataset\cite{11intuitive} comprising EEG recordings from ten participants (7 males and 3 females, right-handed, aged 24-31 years). 
The experimental setup was designed to capture EEG signals across both ME and MI sessions. Each trial began with a 4-second rest period, followed by a 3-second instruction phase, and a 4-second task performance period based on visual cues indicating one of three upper limb movements: forearm extension, hand grasp, or wrist supination. For the ME session, participants physically performed the movement tasks, while in the MI session, they were instructed to vividly imagine the movement without any physical execution. Each task was randomized across 50 trials, culminating in a total of 150 trials per session.

\subsection{Evaluation Metrics} 
In assessing the performance of our multitask learning framework, we computed the classification accuracy and the harmonic mean of precision and recall, commonly known as the F1 score, for both MI and ME tasks. Table \ref{tab:results} shows the performance evaluation results averaged across all participants. We compared our proposed method against the unsparsified baseline model and two established benchmarks for static sparse training, namely the Lottery Ticket Hypothesis (LTH)\cite{lth} and Single-shot Network Pruning (SNIP)\cite{snip}.

\begin{table}[t]
\label{tab:results}
\resizebox{\columnwidth}{!}{%
\begin{tabular}{@{}cccccc@{}}
\toprule
\multirow{2}{*}{Method} &
  \multirow{2}{*}{\begin{tabular}[c]{@{}c@{}}Sparsity\\ (\%)\end{tabular}} &
  \multicolumn{2}{c}{Motor Imagery Task} &
  \multicolumn{2}{c}{Motor Execution Task} \\ \cmidrule(l){3-6} 
 &
   &
  Acc. (\%) &
  F1  &
  Acc. (\%) &
  F1 \\ \midrule
\begin{tabular}[c]{@{}c@{}}Baseline\end{tabular} &
  0 &
  60.0 $\pm$ 4.0 &
  0.59 $\pm$ 0.04 &
  70.0 $\pm$ 3.0 &
  0.69 $\pm$ 0.03 \\ \midrule
               & 20 & 61.0 $\pm$ 4.0 & 0.60 $\pm$ 0.04 & 71.0 $\pm$ 3.0 & 0.70 $\pm$ 0.03 \\
LTH\cite{lth}            & 40 & 62.5 $\pm$ 3.5 & 0.62 $\pm$ 0.03 & 72.5 $\pm$ 2.5 & 0.72 $\pm$ 0.02 \\
               & 80 & 58.0 $\pm$ 5.0 & 0.57 $\pm$ 0.05 & 68.0 $\pm$ 4.0 & 0.67 $\pm$ 0.04 \\ \midrule
               & 20 & 61.5 $\pm$ 3.5 & 0.61 $\pm$ 0.03 & 71.5 $\pm$ 2.5 & 0.71 $\pm$ 0.02 \\
SNIP\cite{snip}           & 40 & 63.0 $\pm$ 3.0 & 0.62 $\pm$ 0.03 & 73.0 $\pm$ 2.0 & 0.72 $\pm$ 0.02 \\
               & 80 & 57.5 $\pm$ 4.5 & 0.57 $\pm$ 0.04 & 67.5 $\pm$ 3.5 & 0.67 $\pm$ 0.03 \\ \midrule
               & 20 & 62.0 $\pm$ 3.0 & 0.61 $\pm$ 0.03 & 72.0 $\pm$ 2.0 & 0.71 $\pm$ 0.02 \\
Ours  & 40 & 64.5 $\pm$ 2.5 & 0.64 $\pm$ 0.02 & 74.5 $\pm$ 1.5 & 0.74 $\pm$ 0.01 \\
               & 80 & 60.5 $\pm$ 4.0 & 0.60 $\pm$ 0.04 & 70.5 $\pm$ 3.0 & 0.70 $\pm$ 0.03 \\ \midrule
\end{tabular}
}
\footnotesize{{$^*$Acc.: accuracy, F1: F1-score}}
\end{table}

\section{RESULTS AND DISCUSSION}
The observed results from the comparative study on sparsity methods suggest a nuanced relationship between sparsity levels and performance metrics across MI and ME tasks. Consistently, the methods displayed a trend where moderate sparsity often improved performance over the unsparsified baseline, which may be attributed to the regularization effect of pruning. This effect likely helps in mitigating overfitting, thereby enhancing the model's generalization capability on unseen data.
Interestingly, our sparse training method demonstrated a robustness to increased sparsity levels, outperforming other methods at higher sparsity percentages. This robustness can be linked to the fact that it is precisely designed to preserve critical network connections that are essential for the tasks at hand, thus maintaining high performance even with a reduced number of parameters. It is also worth noting that the performance in ME tasks consistently surpassed that in MI tasks. This could be due to the more pronounced and consistent signal patterns inherent to executed movements as opposed to imagined ones, which are known to be subtler and more variable. The F1 scores, following a similar trend to accuracy, reinforce the notion that precision and recall are not significantly compromised at moderate levels of sparsity. This balance is crucial in applications where the trade-off between false positives and false negatives is delicate, particularly for BCI systems.

\begin{figure}[!t] 
\centerline{\includegraphics[width=\columnwidth]{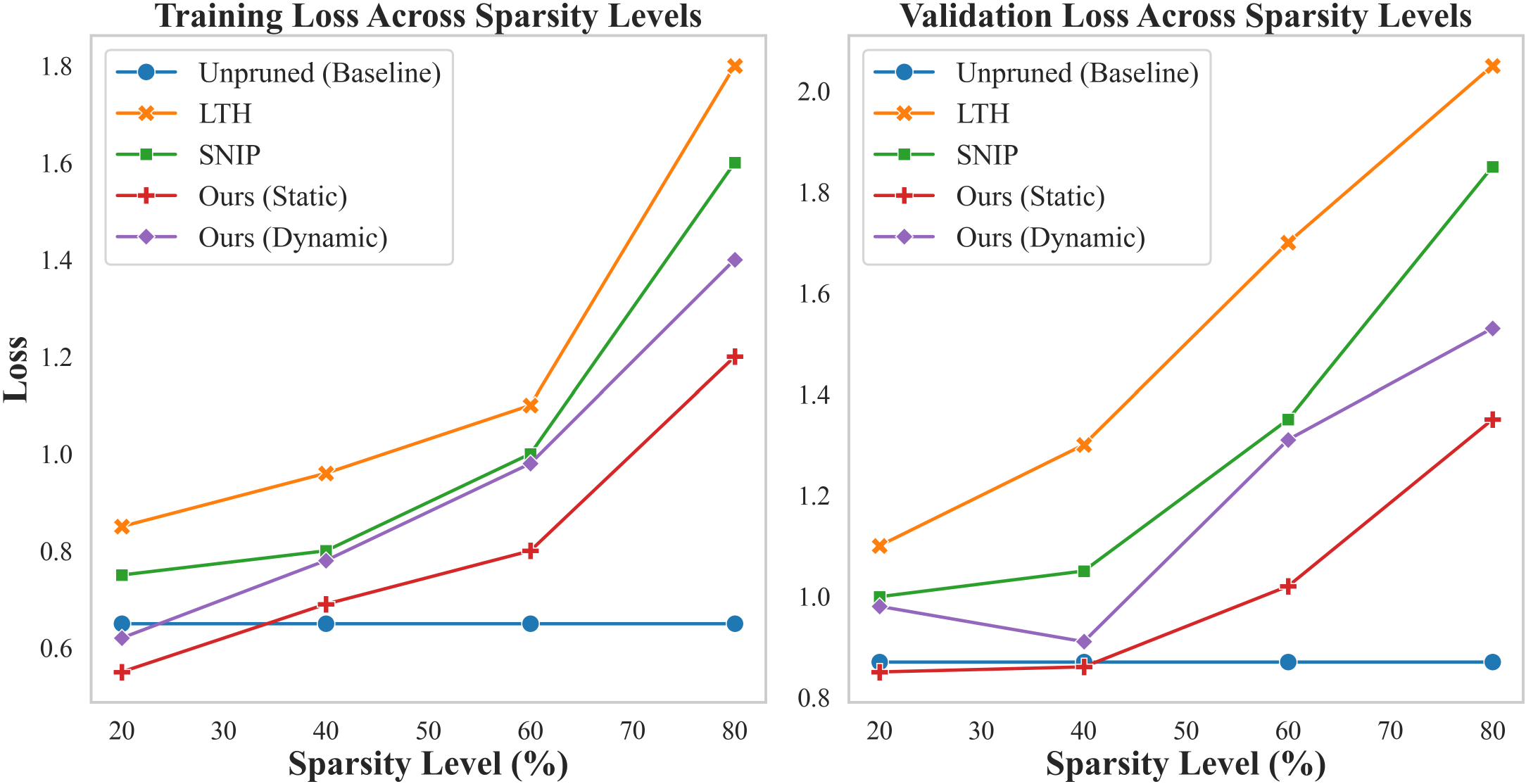}} 
\caption{Convergence of training and validation losses across sparsity levels. The proposed method demonstrates superior convergence rates and lower loss values when compared with alternative methods. This suggests enhanced learning efficiency and model generalization facilitated by our proposed sparse training strategy.}
\label{fig}
\end{figure}

However, as expected, an excessive sparsity level leads to a degradation in performance, likely due to the loss of too many relevant connections, which are critical for capturing the complex patterns required for accurate classification. This performance drop at high sparsity levels serves as a reminder of the delicate balance required in pruning methods to retain essential network functionality.
The observed trends also shed light on the importance of task-specific pruning strategies in multitask learning environments. The distinct performance profiles between MI and ME tasks suggest that a one-size-fits-all approach to sparsity may not be optimal, and task-specific considerations are imperative.
In summary, the findings underscore the potential of tailored sparsity methods in BCI systems, where computational efficiency must be harmonized with robust performance across varied tasks. Future work should explore the integration of these findings into real-world applications and investigate the neurophysiological correlates that underpin the differential performance between MI and ME tasks.

\section{CONCLUSION}
This study presents a novel integration of multitask learning and neural network pruning, specifically applied to motor imagery and execution tasks in EEG-based BCIs. The proposed sparse multitask learning framework not only mitigates the challenge of overfitting but also preserves—and in some instances, enhances—the model's performance even with a reduced parameter set. The results indicate that carefully tailored sparsity can lead to more efficient models without compromising the accuracy and robustness necessary for BCI applications. 
Looking forward, the potential of this methodology extends beyond the current dataset to a broader range of BCI paradigms involving both imagined and executed tasks, such as imagined versus overt speech. The adaptability and effectiveness of our approach in these varied contexts could further validate its utility and robustness. 
Future work will focus on applying our sparse multitask learning framework to diverse BCI datasets and investigating the feasibility of deploying these models in real-time BCI applications. Ultimately, our goal is to bridge the gap between theoretical research and real-world scenarios, paving the way for more robust and efficient BCI systems.

\bibliographystyle{IEEEtran}
\bibliography{REFERENCE}

\end{document}